\pdfoutput=1
\documentclass[letterpaper]{article} % DO NOT CHANGE THIS
\usepackage{aaai2026}  % DO NOT CHANGE THIS
\usepackage{times}  % DO NOT CHANGE THIS
\usepackage{helvet}  % DO NOT CHANGE THIS
\usepackage{courier}  % DO NOT CHANGE THIS
\usepackage[hyphens]{url}  % DO NOT CHANGE THIS
\usepackage{graphicx} % DO NOT CHANGE THIS
\usepackage{natbib}  % DO NOT CHANGE THIS AND DO NOT ADD ANY OPTIONS TO IT
\usepackage{caption} % DO NOT CHANGE THIS AND DO NOT ADD ANY OPTIONS TO IT
\usepackage{algorithm}
\usepackage{algorithmic}
\usepackage{colortbl}
\usepackage{color}
\usepackage[skins,breakable,listings]{tcolorbox}
\tcbuselibrary{listings,theorems}
\definecolor{purple1}{HTML}{8d3a94}
\newtcolorbox{mybox}{colback=white!5!white,colframe=black!75!black, left=.05in, right=.05in}
\newtcbtheorem[]{exmp}{Prompt}%
{colback=purple1!5,colframe=purple1!80,fonttitle=\bfseries, left=.02in, right=.02in,bottom=.02in, top=.02in}{exmp}
\newtcbtheorem[]{exmpa}{Detailed Prompt}%
{enhanced, breakable,colback=purple1!5,colframe=purple1!80,fonttitle=\bfseries, left=.02in, right=.02in,bottom=.02in, top=.02in}{exmp1}
\newtcbtheorem[]{exmpb}{Thinking Process}%
{enhanced, breakable,colback=purple1!5,colframe=purple1!80,fonttitle=\bfseries, left=.02in, right=.02in,bottom=.02in, top=.02in}{exmp2}
\usepackage{newfloat}
\usepackage{listings}
\DeclareCaptionStyle{ruled}{labelfont=normalfont,labelsep=colon,strut=off} % DO NOT CHANGE THIS
\floatstyle{ruled}
\newfloat{listing}{tb}{lst}{}
\floatname{listing}{Listing}
\usepackage{xcolor}
\usepackage{cite}
\usepackage{amsmath}
\usepackage{amsfonts}
\usepackage{amssymb}
\usepackage[caption=false]{subfig}
\usepackage{booktabs}

\newcommand{\blue}[1]{\textcolor{blue}{#1}}
\newcommand{\red}[1]{\textcolor{red}{#1}}

\nocopyright

\title{Vision–Language-Model-Guided Differentiable Ray Tracing for Fast and Accurate Multi-Material RF Parameter Estimation}
\author{
    Zerui Kang\textsuperscript{\rm 1}, Yishen Lim\textsuperscript{\rm 1}, Zhouyou Gu\textsuperscript{\rm 1}, Seung-Woo Ko\textsuperscript{\rm 2}\footnotemark[1], Tony Q.S. Quek\textsuperscript{\rm 1}, Jihong Park\textsuperscript{\rm 1}\thanks{Corresponding authors}
}
\affiliations{
    \textsuperscript{\rm 1}Dept. of Information Systems Technology and Design, Singapore University of Technology and Design\\
    \textsuperscript{\rm 2}Dept. of Smart Mobility Engineering, Inha University\\
    zerui\_kang@mymail.sutd.edu.sg, \{yishen\_lim, zhouyou\_gu, tonyquek, jihong\_park\}@sutd.edu.sg, swko@inha.ac.kr\\ 
}
\usepackage{bibentry}

\begin{document}

\maketitle

\begin{abstract}
Accurate radio-frequency (RF) material parameters are essential for electromagnetic digital twins in 6G systems, yet gradient-based inverse ray tracing (RT) remains sensitive to initialization and costly under limited measurements. This paper proposes a vision–language-model (VLM) guided framework that accelerates and stabilizes multi-material parameter estimation in a differentiable RT (DRT) engine. A VLM parses scene images to infer material categories and maps them to quantitative priors via an ITU-R material table, yielding informed conductivity initializations. The VLM further selects informative transmitter/receiver placements that promote diverse, material-discriminative paths. Starting from these priors, the DRT performs gradient-based refinement using measured received signal strengths. Experiments in NVIDIA Sionna on indoor scenes show 2–4× faster convergence and 10–100× lower final parameter error compared with uniform or random initialization and random placement baselines, achieving sub-0.1\% mean relative error with only a few receivers. Complexity analyses indicate per-iteration time scales near-linearly with the number of materials and measurement setups, while VLM-guided placement reduces the measurements required for accurate recovery. Ablations over RT depth and ray counts confirm further accuracy gains without significant per-iteration overhead. Results demonstrate that semantic priors from VLMs effectively guide physics-based optimization for fast and reliable RF material estimation.
\end{abstract}

\section{Introduction}

Constructing an electromagnetic digital twin is a key enabler for 6G, as it allows the generation of large volumes of physically grounded synthetic channel data to train AI models for site-specific communication tasks such as beamforming, sensing, and resource allocation in smart factories or remote control ~\citep{shu2025digital}. Unlike an optics-focused digital twin that primarily reconstructs geographic shapes ~\citep{tao2022digital}, an electromagnetic digital twin should capture \emph{radio-frequency material} properties, e.g., conductivity and permittivity, because these parameters fundamentally determine multi-path propagation behaviors such as reflection, diffusion, and penetration loss. In this work, given the geometric layout of a scene, we target to estimate the RF material properties in the environment precisely.

A natural route is to cast this as an inverse ray tracing (RT) problem \citep{hoydis2024learning}. RT is referred to as a forward process, producing channel-relevant measurements given geometric and material properties. The inverse task, inferring material properties from measured channel data, is thus well-posed in principle. However, most conventional RT simulators comprise multiple discrete mapping functions (e.g., path enumeration and bounce selection), which make the forward process non-differentiable. Consequently, inverse inference should rely on exhaustive search or heuristic optimization. In contrast, recent differentiable RT engines, such as NVIDIA Sionna \citep{hoydis2022sionna}, provide gradients with respect to the scene parameters. This enables the inverse RT problem to be formulated as a continuous optimization, where material properties can be iteratively refined via gradient-based updates \citep{vaara2025differentiable} rather than a brute-force search. 

Despite these advances, gradient-based inverse RT often fails to reliably estimate material parameters in practice, particularly when multiple materials coexist in the same scene. The key technical challenges are summarized below. First, although differentiable, the inverse problem remains highly sensitive to the initial guess. When the initial material parameters deviate substantially from the ground truth, the iterative refinement may drift toward incorrect parameter regions or require a huge number of updates. This sensitivity makes gradient-based refinement impractical under the tight latency requirements expected for the rapid deployment of 6G systems.
Second, increasing the number of measurements generally improves estimation accuracy, but at a substantial computational overhead. 
Also, materials can hardly be distinguished if the measurements contain insufficient information on the scene. Therefore, the number and positions of measurements should be carefully designed to achieve accuracy and low complexity in the inverse RT.

To address the challenges above, we propose leveraging a vision–language model (VLM) \citep{zhang2024vision} that can capture high-level scene semantics. Specifically, a VLM can parse a scene photograph to infer likely material categories, which serve as informed initialization for the inverse RT process. These material categories can then be mapped to quantitative electromagnetic parameters using an external knowledge base, such as the ITU material tables. In addition, the VLM can suggest a small set of measurement positions that are most informative for material estimation. Incorporating these two VLM-derived priors, namely material parameter initialization and measurement position selections, significantly improves both the accuracy and convergence speed of the inverse RT process.
The simulation results show that our proposed method accelerates the convergence by 2 to 4 times and achieve 10 to 100 times smaller estimation errors of the RF material parameters at convergence.

% , reducing the reliance on brute-force search and mitigating non-convexity effects.   

\section{Related Work}\label{Sec: Related Work}
\subsubsection{RT for Wireless Networks}
The application of RT engines is rapidly expanding beyond simple parameter calculation into complex, system-level tasks. In many applications, these tools are utilized as high-fidelity simulators to generate realistic data or model system performance. For example, they create high-fidelity digital network twins for 6G network simulations \citep{pegurri2025toward} and enable novel sensing applications such as privacy-preserving robot navigation \citep{amatare2024testbed} and real-time object localization \citep{amatare2024real}. In these approaches, radio-frequency (RF) propagation maps generated by the RT engine are used to identify objects \citep{amatare2025rf}, including those in non-line-of-sight (NLoS) conditions. The RT engine is also being explored for simulating reconfigurable intelligent surfaces (RIS) in complex urban environments \citep{gunecser2025ris}.
A more recent trend leverages the differentiability of the RT engine, enabling backpropagation through the propagation model itself and allowing gradient-based optimization. For instance, the differentiable RT engine described in \citep{10705152} provides examples of gradient-based calibration of material properties, scattering patterns, and antenna patterns. Building on this idea, \citet{vaara2025differentiable} propose learning RF material properties (e.g., permittivity, conductivity) to match ground-truth channel impulse responses using point-cloud representations of the scene. Although these works implement gradient-based inverse RT processes, they do not study how to improve the convergence of such processes.

\subsubsection{VLM for Wireless Networks}

Recently, VLMs have also found applications in wireless systems. \citet{zhao2025zero} introduce a training-free framework that utilizes a VLM to identify signal modulation schemes without training data. This approach first converts raw radio signals into various image representations. The VLM then performs reasoning by comparing these images against expert-defined semantic prototypes for each modulation type to determine the best-matched modulation.
Additionally, \citet{wang2025vision} propose a VLM-based multimodal fusion framework for optimal beam prediction in vehicular communications. The framework fuses data from multiple in-vehicle sensors and uses contrastive learning to align semantic embeddings from RGB cameras and LiDAR data. It also incorporates GPS coordinates into a prompt, providing spatial context for the vehicle’s relative location with respect to base stations. The fused embeddings are then fed into a classifier to predict the optimal beam index. These works demonstrate that VLMs can provide rich semantic context from images and prompts to augment wireless tasks without relying on explicit training data.

In summary, it remains an open question how VLMs can be used to exploit the semantic context of a scene to improve the convergence accuracy and speed of the inverse RT process has not yet been studied.

\section{Inverse RT for RF Material Parameter Estimation}
\subsection{RF Material Parameter Estimation}
Our objective is to solve an inverse RT problem that estimates the RF material properties from a limited number of wireless channel measurements within a known geometric scene. The scene has $K$ objects with known geometry as
\begin{align}
\boldsymbol{\Sigma} = \{ \Sigma^{(1)}, \Sigma^{(2)}, \dots, \Sigma^{(K)}\},
\end{align}
where $\Sigma^{(k)}$ denotes the geometry of the $k$-th object. Each object is made of a material with an unknown conductivity parameter collected in
\begin{align}
\boldsymbol{\sigma} = [\sigma^{(1)}, \sigma^{(2)}, \dots, \sigma^{(K)} ]^{\top} \in \mathbb{R}^{K\times 1},
\end{align}
where $\sigma^{(k)}$ denotes the unknown conductivity of the $k$-th object in Siemens/meter, i.e., the parameter to be estimated\footnote{Each object's relative permittivity, also an important RT parameter, is assumed to be given for simplicity. Extending the formulation to jointly estimate conductivity and permittivity is conceptually straightforward and remains future work.}.  

We consider that $M$ measurement trials are made in the scene, each using one transmitter and multiple receivers. In the $m$-th trial, the transmitter is located at $\mathbf{x}_m \in \mathbb{R}^{3\times 1}$ and $N$ receivers are placed at $\mathbf{y}_{m,1},\dots, \mathbf{y}_{m,N}$, where $\mathbf{y}_{m,n}\in \mathbb{R}^{3\times 1}$ denotes the position of the $n$-th receiver. We define the full measurement configuration as
\begin{align}
\mathbf{P}_m= [\mathbf{x}_m, \mathbf{y}_{m,1}, \dots, \mathbf{y}_{m,N}]\in \mathbb{R}^{3\times (1+N)}, m=1,\dots,M.
\end{align}
The corresponding ground-truth received signal strengths are collected in
\begin{align}
{\mathbf{r}}_m \triangleq [{r}_{m,1}, \dots, {r}_{m,N}]\in \mathbb{R}^{1\times N},
\end{align}
where ${r}_{m,n}$ is the received signal strength at the $n$-th receiver in the $m$-th trial.  

To estimate the material parameters, we construct a digital twin that models the scene geometry $\boldsymbol{\Sigma}$ and associates each object with the estimated conductivity parameter.
We employ the differentiable RT engine, e.g., NVIDIA Sionna, to simulate the wireless channel response. The engine operates at a carrier frequency $f_{\text{c}}$ and launches $U_{\text{ray}}$ rays from the transmitter with a transmission power $P_{\text{Tx}}$. 
These rays propagate through the digital twin, interacting with objects up to $D$ times (the ray-tracing depth), with each interaction determined by the material conductivity $\boldsymbol{\sigma}$. The simulation parameters in the engine are configured as
\begin{equation}
\mathcal{C} = \{f_{\text{c}}, P^{\text{Tx}}, G^\text{Tx}, G^\text{Rx}, U_{\text{ray}}, D\},
\end{equation}
where $P^{\text{Tx}}$ is the transmission power, and $G^\text{Tx}$, $G^\text{Rx}$ are the antenna patterns of the transmitter and receivers, respectively.
For each measurement trial $m\in\{1,\dots,M\}$, the engine simulates the received signal strength as
\begin{equation}
\mathbf{r}^\mathrm{RT}_m\triangleq [r^\mathrm{RT}_{m,1},\dots,r^\mathrm{RT}_{m,N}]^{\top} = \mathrm{RT}(\boldsymbol{\sigma},\mathbf{P}_m,\boldsymbol{\Sigma}, \mathcal{C}),
\end{equation}
where $r^\mathrm{RT}_{m,n}$ denotes the simulated signal strength at receiver location $\mathbf{y}_{m,n}$, and $\mathrm{RT}(\cdot)$ denotes the differentiable RT engine simulation.
The material parameter estimation can now be formulated as solving a system of $M$ nonlinear equations
\begin{equation}\label{eq:system_of_equation}
\mathbf{r}_m = \mathbf{r}^\mathrm{RT}_m|_{\mathbf{r}^\mathrm{RT}_m=\mathrm{RT}(\boldsymbol{\sigma},\mathbf{P}_m,\boldsymbol{\Sigma}, \mathcal{C})}, \forall m\in\{1,\dots,M\}.
\end{equation}
This system is highly nonlinear due to the complex ray interactions within the differentiable RT engine, where conductivity parameters $\boldsymbol{\sigma}$ affects multiple rays through reflections, refractions, and attenuations across all measurement trials.

\subsection{Gradient-Based Inverse RT Process}
The above system of nonlinear equations in \eqref{eq:system_of_equation} is solved by finding the parameter vector ${\boldsymbol{\sigma}}$ that minimizes the difference between the simulated signal strengths $\mathbf{r}^\mathrm{RT}_m$ and the ground-truth signal strengths $\mathbf{r}_m$, $m=1,\dots,M$.
Utilizing the differentiability of the RT engine with respect to the conductivity parameters $\boldsymbol{\sigma}$, we define a loss function that captures this difference between simulated and measured signal strength and minimize it using the gradient-based optimization via backpropagation through the differentiable RT computation graph.
Specifically, the loss function on the difference is defined over all $M$ measurement trials, e.g., using the mean normalized absolute error (NAE), as
\begin{equation}\label{eq:nae_loss}
\begin{aligned}
    &\mathcal{L}(\boldsymbol{\sigma}|\mathbf{P}_m,\boldsymbol{\Sigma}, \mathcal{C}) \\
    =&\sum_{m=1}^{M}\frac{1}{N} \sum_{n=1}^{N} \frac{\left| {r}_{m,n} - r^\mathrm{RT}_{m,n}\right|}{{r}_{m,n}}\Big|_{\mathbf{r}_m = \mathrm{RT}(\boldsymbol{\sigma}, \mathbf{P}_m,\boldsymbol{\Sigma}, \mathcal{C})}.
\end{aligned}
\end{equation}
Then, the optimal RF material parameter estimation is equivalent to solving the loss minimization problem as
\begin{equation}
\min_{\boldsymbol{\sigma}} \mathcal{L}(\boldsymbol{\sigma}|\mathbf{P}_m,\boldsymbol{\Sigma}, \mathcal{C}).
\end{equation}
Leveraging the differentiability, we can efficiently compute the gradient of the loss function $\mathcal{L}$ with respect to the conductivity parameters $\boldsymbol{\sigma}$ as $\nabla_{\boldsymbol{\sigma}} \mathcal{L}({\boldsymbol{\sigma}})$, and iteratively update the parameters to minimize the loss via gradient descent as
\begin{equation}\label{eq:backpropagation}
\boldsymbol{\sigma}(i+1) = \boldsymbol{\sigma}(i) - \eta \nabla_{\boldsymbol{\sigma}(i)} \mathcal{L}\big(\boldsymbol{\sigma}(i)\big), \ i=1,2,3,\dots,
\end{equation}
where $\eta$ is the learning rate and $\boldsymbol{\sigma}(i)$ is the estimated conductivity parameters at the $i$-th iteration. The initial values of the conductivities are
\begin{equation}\label{eq:parameter_init}
\boldsymbol{\sigma}(1) = \boldsymbol{\sigma}^{\text{init}}.
\end{equation}
Solving the above gradient-based inverse problem raises two main challenges: 1) improving the accuracy of the estimated parameters by ensuring their convergence to the ground-truth values and 2) reducing the computational complexity to achieve a short total time to convergence. 

To better understand the complexity of the iterative gradient-based inverse RT process, we analyze the complexity with respect to the number $K$ of objects/materials and the number $M$ of measurement trials when $N=3$ receivers are used. We measure 1) the scene building time, i.e., time spent on loading the geometry $\boldsymbol{\Sigma}$,  2) the average ray-tracing forward propagation time per iteration and 3) the backpropagation time per iteration.
\begin{figure}[t] 
\centering 
\includegraphics[width=0.9\columnwidth]{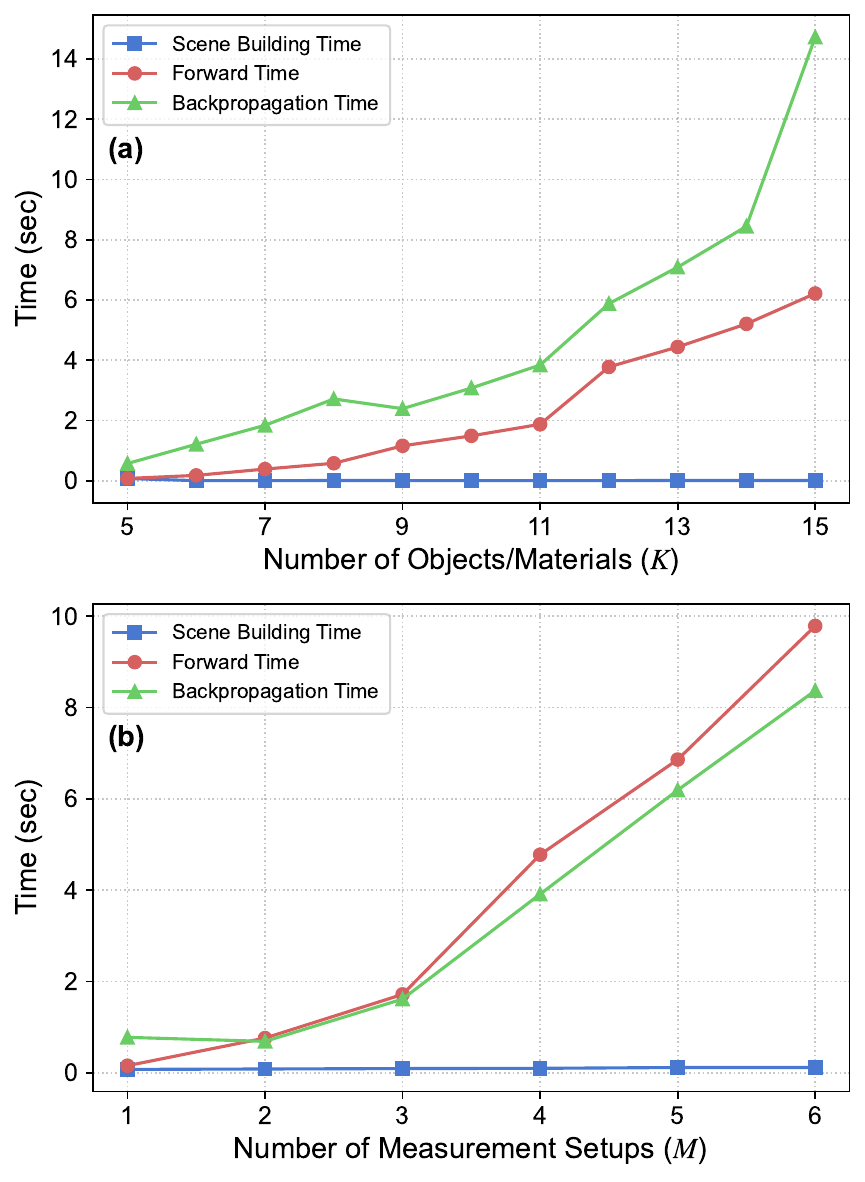}
\caption{Computing time per iteration of the RT engine.}
\label{fig:scalability_plots} 
\vspace{-0.2cm}
\end{figure}

When analyzing the complexity with respect to the number $K$ of objects/materials, we vary $K$ from $5$ to $15$ by inserting additional boxes with different materials into the scene and distributing them evenly in the room, while fixing $M=1$. As shown in Fig.~\ref{fig:scalability_plots}a, the scene building time remains nearly constant as $K$ increases, since geometry loading incurs only a small overhead compared to the ray tracing computations. In contrast, the forward-propagation time of the RT engine grows approximately linearly with $K$, reflecting the increased number of ray–object interactions. The backpropagation time grows even faster, at a super-linear rate, as the underlying computation graph becomes more complex and its gradient paths expand with more objects. This super-linear scaling with the number of objects $K$ presents a significant computational bottleneck, making the inverse problem intractable for complex, real-world scenes with many materials.

We also study the complexity with respect to the number of measurement setups $M$, shown in Fig.~\ref{fig:scalability_plots}b. The scene building time again stays negligible for all values of $M$, while both forward and backpropagation times increase roughly linearly as more measurement setups are added. This linear growth highlights the importance of limiting the number of measurement trials in order to maintain a low per-iteration computational cost and ensure practical convergence times. This creates a difficult trade-off: while more measurements $M$ are generally needed to ensure accuracy and convergence to the ground-truth values, each additional measurement setup linearly increases the per-iteration computational cost, leading to prohibitively long convergence times.

Based on this complexity analysis, we next propose a VLM-guided RF material parameter estimation framework that intelligently initializes the conductivity parameters and selects informative measurement positions to accelerate convergence.

\section{VLM-Guided Inverse RT for RF Material Parameter Estimation}
This section presents our proposed VLM-guided framework that addresses the convergence accuracy and complexity challenges inherent in differentiable RT-based material parameter estimation.
We first formulate the optimization of the inverse RT process by initializing the conductivity parameters and selecting measurement positions. We then detail the VLM-aided initialization and position selection methods with specific prompt designs, and finally summarize the overall procedure.

\subsection{Optimization of the Inverse RT Process}
We first formulate the optimization of the inverse RT process, aiming to minimize the total computation time while ensuring convergence to accurate RF material parameters.
Each iteration of the gradient-descent \eqref{eq:backpropagation} is assumed to cost approximately the same time $\Delta (N,M,\boldsymbol{\Sigma},\mathcal{C})$ due to the fixed scene setup \citep{hoydis2022sionna}: the number $N$ of measurements per trial, the number $M$ of measurement trials, the geometry $\boldsymbol{\Sigma}$, and the RT configurations $\mathcal{C}$.
The iterations stop when a convergence criterion is satisfied.
Ideally, the convergence at the $I$-th iteration occurs when the iterated parameters $\boldsymbol{\sigma}(i)$ are sufficiently close to the ground-truth ones $\hat{\boldsymbol{\sigma}}\triangleq[\hat{\sigma}^{(1)}, \hat{\sigma}^{(2)}, \dots, \hat{\sigma}^{(K)} ]^{\top}$ as
\begin{equation}\label{eq:ideal_stopping_criterion}
    \boldsymbol{\sigma}(I) = \hat{\boldsymbol{\sigma}}.
\end{equation}
However, since the ground-truth parameters $\hat{\boldsymbol{\sigma}}$ are unknown in practice, we instead define the convergence based on the stability of the loss function and the estimated parameters over the last $J$ iterations as
\begin{equation}\label{eq:stopping_criterion}
    \begin{aligned}
        |\mathcal{L}\left(\boldsymbol{\sigma}(i)\right) - \mathcal{L}\left(\boldsymbol{\sigma}(i+1)\right)|&\leq \alpha,\ \forall i=I-J,\dots,I,\\
        |\sigma^{(k)}(i)-\sigma^{(k)}(i+1)| &\leq \beta,\ \forall k, \forall i=I-J,\dots,I,
    \end{aligned}
\end{equation}
where $\alpha$ and $\beta$ are small tolerances, and $J$ is the patience parameter.
The total computation time for convergence can thus be approximated as $I\Delta(N,M,\boldsymbol{\Sigma},\mathcal{C})$ where $I$ is the minimum iteration index satisfying \eqref{eq:stopping_criterion}.

To reduce this total computation time while ensuring convergence, we observe that both the initialization of the conductivity parameters $\boldsymbol{\sigma}^{\text{init}}$ and the measurement positions $\mathbf{P}_m$ $\forall m$ significantly influence the number of iterations $I$ required for convergence. Specifically, as shown in \eqref{eq:backpropagation}, the gradient descent iteratively updates the conductivities starting from initial values $\boldsymbol{\sigma}^{\text{init}}$. Therefore, a good initialization can place the starting point in a well-conditioned region of the loss landscape, reducing the number of iterations required for convergence. Also, the gradient values depend on the measurement positions $\mathbf{P}_m$ $\forall m$, which determine the rays' interactions with different materials in the scene. Choosing informative measurement positions can enhance the sensitivity of the loss function to changes in the conductivity parameters, leading to more effective updates. Moreover, effective measurement positions can reduce the number $N$ of measurements per trial, and thus further decrease the computation time per iteration.
The optimization problem is thus to jointly optimize the initial values of the conductivity parameters $\boldsymbol{\sigma}^{\text{init}}$ and the measurement positions $\mathbf{P}_m$ $\forall m$ as 
\begin{equation}
    \label{eq:optimization_problem}
    \begin{aligned}
        \min_{\mathbf{\sigma}^{\text{init}};\ \mathbf{P}_m, m=1,\dots,M}\ I\Delta(N,M,\boldsymbol{\Sigma},\mathcal{C}),\ \ \text{s.t. }\eqref{eq:backpropagation}\eqref{eq:parameter_init}\eqref{eq:stopping_criterion}.
    \end{aligned}
\end{equation}

\begin{figure}[t] 
  \centering
  \includegraphics[width=0.95\columnwidth]{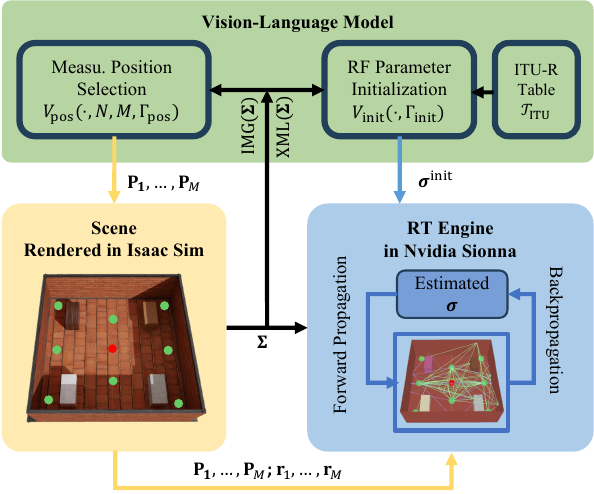} 
  \caption{Illustration of VLM-guided inverse RT Process.}
  \label{fig:framework}
  \vspace{-0.2cm}
\end{figure}

\begin{table}[ht]
    \centering 
    \begin{exmp}{VLM-Aid RF Material Parameter Init.}{}
        \small
        Inputs:

        1. Scene Image: A photograph of the environment.

        2. ITU-R Material Table: A dictionary mapping standard material names to their (c, d) conductivity parameters.

        You are an expert radio frequency (RF) engineer specializing in parameter estimation. Provide intelligent initial guesses for the frequency-dependent conductivity parameters (c, d) for all visible materials in the provided scene image.

        Task:

        1. Identify all objects and their corresponding materials in the Scene Image.

        2. For each material, select the closest match from the ITU-R Material Table and assign its (c, d) parameters. If a direct match is unavailable, use the parameters of the most similar material as a default guess.

        Output Schema: A JSON list of objects with RF parameters.
    \end{exmp}
    \vspace{-0.2cm}
\end{table}

\begin{table}[ht]
    \centering 
    \begin{exmp}{VLM-Aided Measurement Position Sel.}{}
        \small
    Inputs:

    1. Scene Image: An overhead view or a key 3D perspective of the environment.

    2. Geometry XML: The structured XML description of the scene.

    3. Measurement Counts: N (Number of Receivers) and M (Number of measurement configurations).

    You are an intelligent spatial planner optimizing RF measurement campaigns. Your goal is to select Transmitter (Tx) and Receiver (Rx) positions that maximize the diversity and information gain regarding the conductivity parameters of the materials in the scene.

    Task:

    1. Analyze the XML and Image to identify all distinct material interfaces and complex geometrical features.

    2. Select N distinct Tx/Rx positions and form M pairs. Prioritize positions that force the RF path to interact with different materials (e.g., Non-Line-of-Sight (NLoS) or Reflection-dominant paths).

    3. Provide the selected positions as a JSON list. The coordinates must be valid and directly correspond to the XML's coordinate system.

    Output Schema: A JSON list of measurement positions.
    \end{exmp}
    \vspace{-0.2cm}
\end{table}

\subsection{VLM-Guided Inverse RT Process}
Despite the clear formulation, solving the above problem is challenging due to complex interactions between the initialization $\boldsymbol{\sigma}^{\text{init}}$, measurement positions $\mathbf{P}_m$ $\forall m$, and the differentiable RT engine's nonlinear dynamics.
We utilize the VLM to provide high-level semantic priors that guide both the initialization of the conductivity parameters and the selection of measurement positions, thereby accelerating convergence and improving estimation accuracy.

For parameter initialization, we utilize the VLM's ability to extract semantic information from the image of the scene and map them to standard material properties. Let $V_{\text{Init}}$ denote the VLM-based initialization function, which takes the geometry description on $\boldsymbol{\Sigma}$, the images on the scene $\mathrm{IMG}(\boldsymbol{\Sigma})$, the ITU-R material table $\mathcal{T}_{\text{ITU}}$ and the query prompt $\Gamma_\text{init}$ as inputs and outputs the initial conductivities $\boldsymbol{\sigma}^{\text{init}}$ as
\begin{equation}\label{eq:vlm_initialization}
\boldsymbol{\sigma}^{\text{init}} = V_{\text{init}}(\mathrm{IMG}(\boldsymbol{\Sigma}), \mathcal{T}_{\text{ITU}},\Gamma_\text{init}).
\end{equation}
For measurement position selection, we leverage the VLM's spatial reasoning capabilities to select transmitter and receiver positions that maximize information gain. Let $V_{\text{Pos}}$ denote the VLM-based selection function, which takes the image of the scene $\mathrm{IMG}(\boldsymbol{\Sigma})$, the geometry description of $\boldsymbol{\Sigma}$ in XML format as $\mathrm{XML}(\boldsymbol{\Sigma})$ and the query prompt $\Gamma_\text{pos}$ including the numbers of measurements $N$ and $M$ as inputs and outputs the selected measurement positions $\mathbf{P}_m$ $\forall m$ as
\begin{equation}\label{eq:vlm_position_selection}
\{\mathbf{P}_1,\dots, \mathbf{P}_M\} = V_{\text{pos}}(\mathrm{IMG}(\boldsymbol{\Sigma}), \mathrm{XML}(\boldsymbol{\Sigma}),N,M,\Gamma_\text{pos}).
\end{equation}
The final step is to perform gradient-based refinement to solve the non-linear equations using the differentiable RT engine, starting from the initialized parameters and using the selected measurement positions. The optimization iteratively updates the parameters according to \eqref{eq:backpropagation} until the stopping criterion in \eqref{eq:stopping_criterion} is met, yielding the estimated conductivities. The overall procedure is summarized as Algorithm~\ref{alg:vlm_drt} and the prompts used in the VLM functions are summarized in Prompt 1 and Prompt 2 and detailed in the appendix.

\begin{algorithm}[t]
\caption{VLM-Guided Inverse RT Process for RF Material Parameter Estimation}\label{alg:vlm_drt}
\begin{algorithmic}[1]
\STATE \textbf{Input:} Geometry $\boldsymbol{\Sigma}$; image $\mathrm{IMG}(\boldsymbol{\Sigma})$; XML $\mathrm{XML}(\boldsymbol{\Sigma})$; ITU table $\mathcal{T}_{\text{ITU}}$; Sionna engine $\mathrm{RT}(\cdot)$; engine configure $\mathcal{C}$; VLM $V_{\text{init}}$ or $V_{\text{pos}}$; prompts $(\Gamma_{\text{init}}, \Gamma_{\text{pos}})$; learning rate $\eta$; max iter. $I$; tolerances $(\alpha,\beta)$; patience $J$; numbers of measurements $(N,M)$.
\STATE \textbf{Output:} Estimated conductivities $\boldsymbol{\sigma}$.
\STATE VLM initialization on $\boldsymbol{\sigma}^{\text{init}}$ as \eqref{eq:vlm_initialization} and set $\boldsymbol{\sigma}(1) \leftarrow \boldsymbol{\sigma}^{\text{init}}$.
\STATE VLM selection on positions $\mathbf{P}_m$ $\forall m$ as \eqref{eq:vlm_position_selection}.
\STATE Measure ground-truth signal strengths, $\{\mathbf{r}_m\}_{m=1}^M$.
\FOR{$i$=1,2,3,$\dots$}
    \FOR{$m=1$ to $M$}
        \STATE Compute $\mathbf{r}^\mathrm{RT}_m$ in differentiable RT engine.
    \ENDFOR
    \STATE Compute loss $\mathcal{L}(\boldsymbol{\sigma}(i))$ from difference between $\{\mathbf{r}^\mathrm{RT}_m\}_{m=1}^M$ and $\{\mathbf{r}_m\}_{m=1}^M$ as \eqref{eq:nae_loss}.
    \STATE Compute gradient $\nabla_{\boldsymbol{\sigma}(i)} \mathcal{L}(\boldsymbol{\sigma}(i))$.
    \STATE Perform gradient descent update as \eqref{eq:backpropagation}.
    \IF{\eqref{eq:stopping_criterion} is satisfied}
        \STATE $I\leftarrow i$ and \textbf{break}
    \ENDIF
    \STATE $i \leftarrow i+1$.
\ENDFOR
\STATE \textbf{return} $\boldsymbol{\sigma}(I)$ as estimated $\boldsymbol{\sigma}$.
\end{algorithmic}
\end{algorithm}

\section{Evaluations}
\subsection{Experiment Setup}

% \begin{figure}[!t] 
%     \centering 
%     \includegraphics[width=0.5\textwidth]{scene.png} 
%     \caption{The experiment scene with ray tracing showing the 3D geometry $\boldsymbol{\Sigma}$ consisting of $M=9$ objects (one floor, four walls, and four boxes made of different materials). Multiple rays from each transmitter position $\mathbf{p}^\text{Tx}_s$ interact with objects up to $D=6$ times before reaching receivers at positions $\mathbf{p}^\text{Rx}_{s,r}$.}
%     \label{fig:Experiment scene with Ray Tracing.}
% \end{figure}

\subsubsection{System Configurations}\label{sec:setup}
The simulation environment is running on the NVIDIA Sionna differentiable RT engine running and one Intel Core Ultra 9 Processor 285K CPU with one Geforce RTX 5090 GPU. We simulate a $10\,\text{m} \times 10\,\text{m} \times 3\,\text{m}$ indoor room with known geometry $\boldsymbol{\Sigma}$ consisting of $K=9$ objects: one brick floor, four brick walls surrounding the room, and four boxes with different materials, e.g., wood, chipboard, concrete and marble. 
Each box has dimensions 1m $\times$ 2m $\times$ 1m (width $\times$ length $\times$ height). 
The wooden box is centered at (-3m,3m), the concrete box at (3m,-3m), the marble box at (-3m,-3m), and the chipboard box at (3m,3m). More rendered scenes are shown in the appendix.
For the RT configurations $\mathcal{C}$, the system operates at carrier frequency $f_c=3.5$ GHz with transmission power $P^\text{Tx}=44$ dBm of the transmitter.  
Both transmitter and receivers are modeled as single-element planar arrays with an isotropic radiation pattern and vertical polarization. 
The engine launches $U_\text{ray}=5000$ rays from the transmitter to simulate the RF signals with the ray-tracing depth $D=4$.
We vary the number $M$ of measurement trials and the number $N$ of receivers per trial.

We assume that the ground-truth conductivity parameters $\boldsymbol{\sigma}$ are perturbed standard values $\boldsymbol{\sigma}^\text{ITU}$ from the ITU-R P.2040 recommendations \citep{conrat2024material}, as prompted in $\Gamma_\text{init}$ listed in the appendix.
Specifically, we set $\sigma_m = \lambda \cdot \sigma^\text{ITU}_m$ for each material $m$, where $\lambda \sim \mathcal{N}(1.0,\,0.1^2)$ is truncated to $[0.8,\,1.2]$ to enable realistic variations. For each measurement trial $m=1,\dots,M$, we simulate the ground-truth signal strength $\mathbf{r}_m$ at receiver positions $\mathbf{P}_m$ using NVIDIA Sionna with the ground-truth conductivities, which we have discussed previously.
The gradient descent optimization in \eqref{eq:backpropagation} uses the Adam ~\citep{Jakob2020DrJit} optimizer with learning rate $\eta=10^{-3}$.
We restrict the maximum number of iterations to $1000$.
The early stopping criterion in \eqref{eq:stopping_criterion} uses tolerances $\alpha=10^{-5}$, $\beta=10^{-4}$ and patience $J=50$ iterations. We utilize Google Gemini 2.5 Pro ~\citep{comanici2025gemini} as the VLM for both initialization and position selection. The average time to process queries for parameter initialization and measurement position selection is around $3.8$s and $50.1$s, respectively. Since the VLM processing time is much shorter than the iteration time of the inverse RT process and does not vary significantly with the different scene setups, we will focus on the time of inverse RT process in the remainder of the experiments.
% \begin{table}[!t] Bı
%     \centering 
%     \caption{Material Conductivities from ITU-R P.2040.} 
%     \label{tab:itu_materials_conductivity}
%     \begin{tabular}{l cc l} 
%     \toprule
%     & \multicolumn{2}{c}{Conductivity [S/m]} &  \\
%     \cmidrule(lr){2-3} 
%     Material & c & d & Frequency (GHz) \\
%     \midrule
%     concrete & 0.0462 & 0.7822 & 1 – 100 \\
%     brick & 0.0238 & 0.1600 & 1 – 40 \\
%     wood & 0.0047 & 1.0718 & 0.001 – 100 \\
%     chipboard & 0.0217 & 0.7800 & 1 – 100 \\
%     marble & 0.0055 & 0.9262 & 1 – 60 \\
%     \bottomrule
%     \end{tabular}
% \end{table} 

\subsubsection{Baseline Methods}
We refer to our proposed VLM RF material parameter initialization and position selection method as \textbf{VLMInit} and \textbf{VLMSel}, respectively. We compare our scheme with the cases as follows. 
\begin{itemize}
    % \item \textbf{VLMInit \& w/o DRT:} Using only the VLM-predicted parameters without further optimization via backpropagation through differentiable RT.
    \item \textbf{RandInit:} Random initialization with $\boldsymbol{\sigma}^\text{init}$ sampled from a uniform distribution as $\mathcal{U}[0.01, 0.06]$ S/m.
    \item \textbf{UnifInit:} Uniform initialization of $\boldsymbol{\sigma}^\text{init}$ with average values of ITU-R material conductivities \citep{conrat2024material}.
    \item \textbf{RandSel:} Random selection of measurement positions within the scene.
\end{itemize}

\subsubsection{Evaluation Metrics}
We evaluate the accuracy of the estimated conductivities using mean relative error (MRE) between the iterated values and the ground truth in the $i$-th iteration as
\begin{equation}
\text{MRE}(i) = \frac{1}{K} \sum_{k=1}^{K} \frac{|\sigma^{(k)}(i) - \hat{\sigma}^{(k)}|}{\hat{\sigma}^{(k)}}.
\end{equation}
We also measure the number of iterations and the total time spent until convergence.

\subsection{Performance of VLM-Aid Initialization}
% \begin{table}[t]
%     \centering
%     \caption{Comparison of optimization time and estimation accuracy for different initialization methods.}
%     \label{tab:init_performance}
%     \begin{tabular}{l@{\hskip 6pt}c@{\hskip 6pt}c@{\hskip 4pt}c} 
%     \toprule
%     Method & MRE (\%) & Iter. & Time (s) \\
%     \midrule
%     \textbf{VLMInit \& VLMSel} & \textbf{0.0106} & \textbf{312} & \textbf{1097.87} \\
%     UnifInit \& VLMSel        & 0.0109 & 818  & 2862.23 \\
%     RandInit \& VLMSel        & 0.0153 & 1324 & 4583.47 \\
%     VLMInit \& w/oDRT         & 7.4527 & --   & -- \\
%     \bottomrule
%     \end{tabular}
% \end{table}

\begin{figure}[!t] 
    \centering
    \includegraphics[width=0.96\columnwidth]{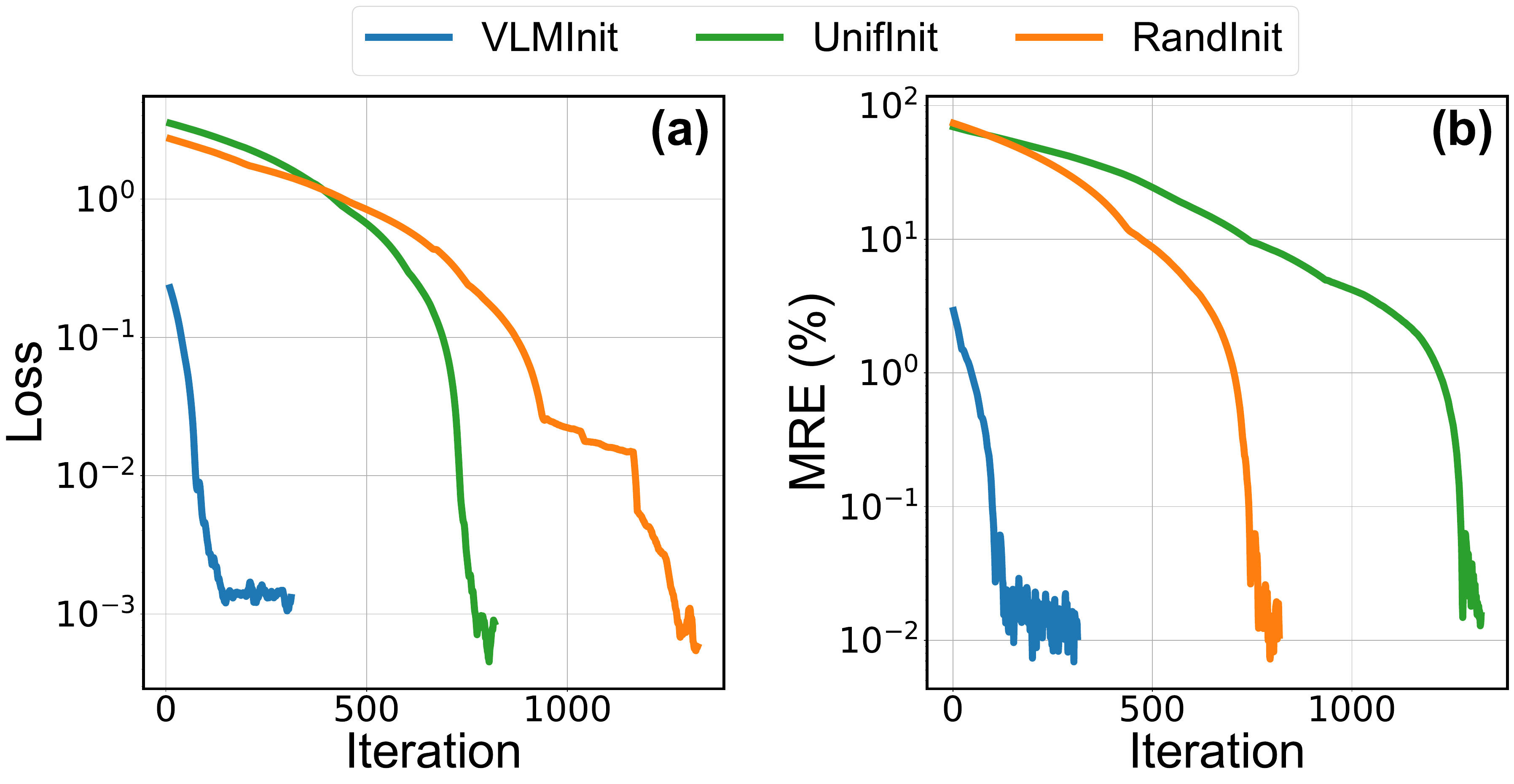}
    \caption{Convergence of different initialization methods in a) loss and b) MRE values over iterations.} 
    \label{fig:threemethod-combined}
    \vspace{-0.2cm}
\end{figure}

\begin{figure}[!t] 
    \centering
    \includegraphics[width=0.96\columnwidth]{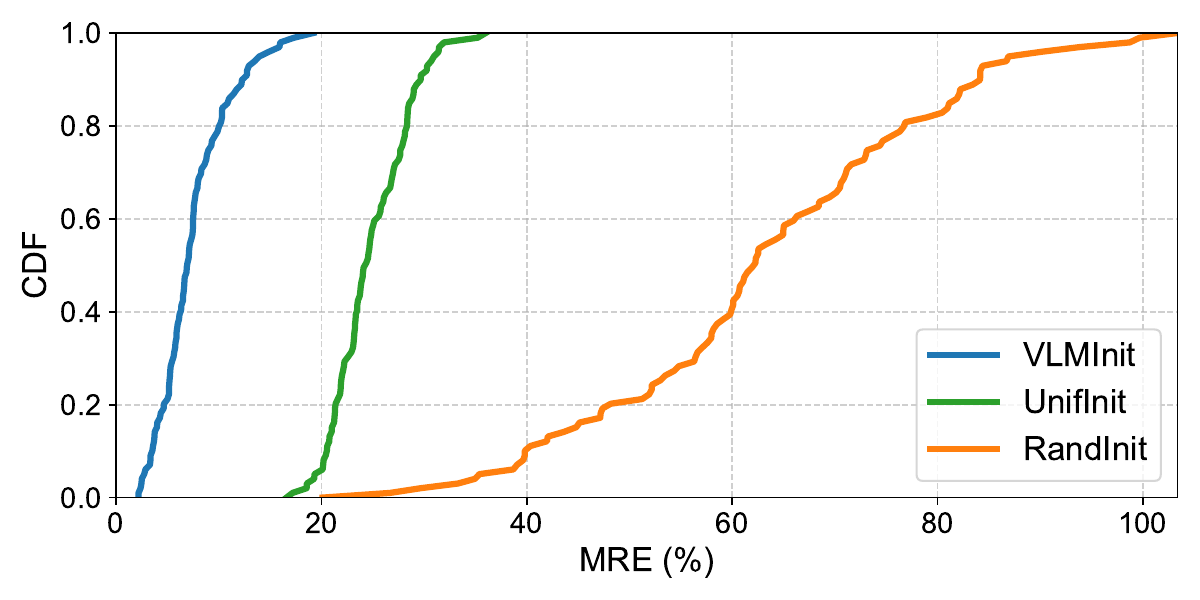}
    \caption{CDF of initial MRE for different initializations.}
    \label{fig:cdf_plot}
    \vspace{-0.2cm}
\end{figure}

\begin{figure}[!ht]
    \centering
    \includegraphics[width=0.96\columnwidth]{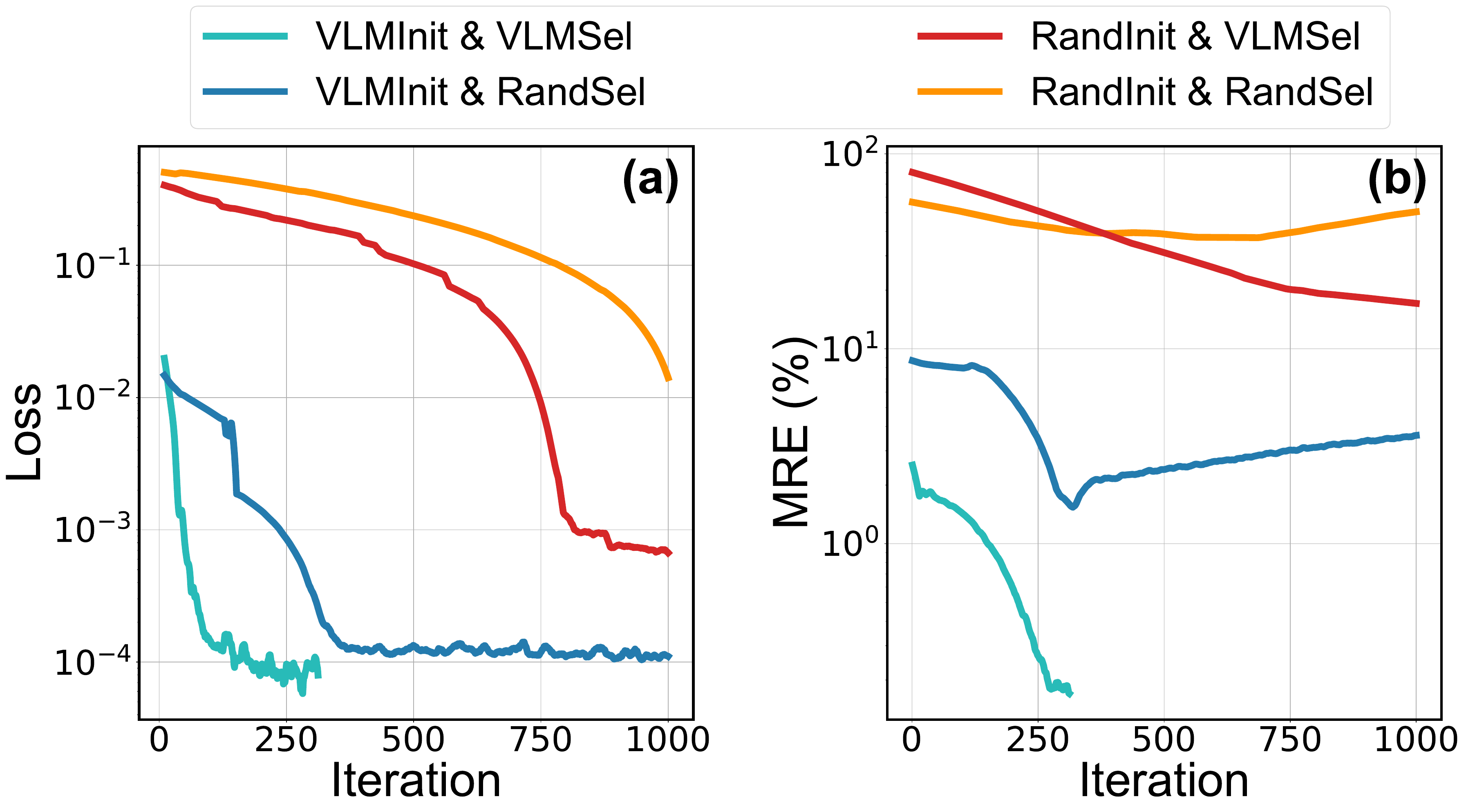}
    \caption{Convergence of different initialization and position selection methods in a) loss and b) MRE values over iterations. }
    \label{fig:my_combined_plot}
    \vspace{-0.2cm}
\end{figure}
To evaluate the VLM initialization on the conductivities, we provide a sufficient amount of measurements in each measurement trial to ensure convergence. Specifically, we let $N=8$ receivers per measurement trial for each of the $M=3$ measurement trials, where the positions are selected by the VLM and are illustrated in the appendix.
To evaluate the effectiveness of our VLM initialization (VLMInit), we compare its performance against three baselines: VLMInit \& w/oDRT,  UnifInit \& VLMSel and RandInit \& VLMSel, as presented in Fig.~\ref{fig:threemethod-combined}. 
Our VLMInit method significantly outperforms the baselines in the convergence speed and the estimation accuracy. Specifically, VLMInit achieves convergence with the fewest number of iterations. In contrast, UnifInit and RandInit require over 2 and 4 times more iterations and time to reach a comparable MRE value.

To further illustrate the accuracy of the initialization provided by the VLM, we analyze its initial estimation quality across 100 samples and present the cumulative distribution function (CDF) in Fig.~\ref{fig:cdf_plot}. 
The results show that the VLMInit achieves a much lower initial MRE value, up to 3 times or 10 times, respectively, than the random and uniform initializations, respectively. This further validates its effectiveness in providing a good starting point for the optimization process, which leads to faster convergence to the ground-truth values.

% When using VLMInit without the gradient-based inverse RT (VLMInit \& w/o DRT), the initial MRE is as high as 7.4527\%, indicating that while the VLM provides a reasonable starting point, it lacks the precision needed for accurate material parameter estimation without further optimization.

% \begin{table}[h]
%     \centering
%     \caption{Comparison of Postion Selection Methods.}
%     \label{tab:strategy_performance}
%     \begin{tabular}{cc@{\hskip 6pt}c@{\hskip 6pt}c}
%     \toprule
%     Method & MRE (\%) & Iter. & Time (s) \\
%     \midrule
%     VLMInit \& VLMSel (R=2)     & 1.770   & 128  & 464.21  \\
%     VLMInit \& VLMSel (R=3)     & 0.170   & 361  & 1097.88 \\
%     VLMInit \& VLMSel (R=4)     & 0.070   & 191  & 703.07  \\
%     VLMInit \& RandSel (R=3)    & 3.597   & 1000 & 3552.32 \\
%     RandInit \& RandSel (R=3)   & 50.433  & 1000 & 3336.52 \\
%     \bottomrule
%     \end{tabular}
% \end{table}

\subsection{Performance of VLM Position Selection}
\begin{figure}[!ht]
    \centering
    \includegraphics[width=0.96\columnwidth]{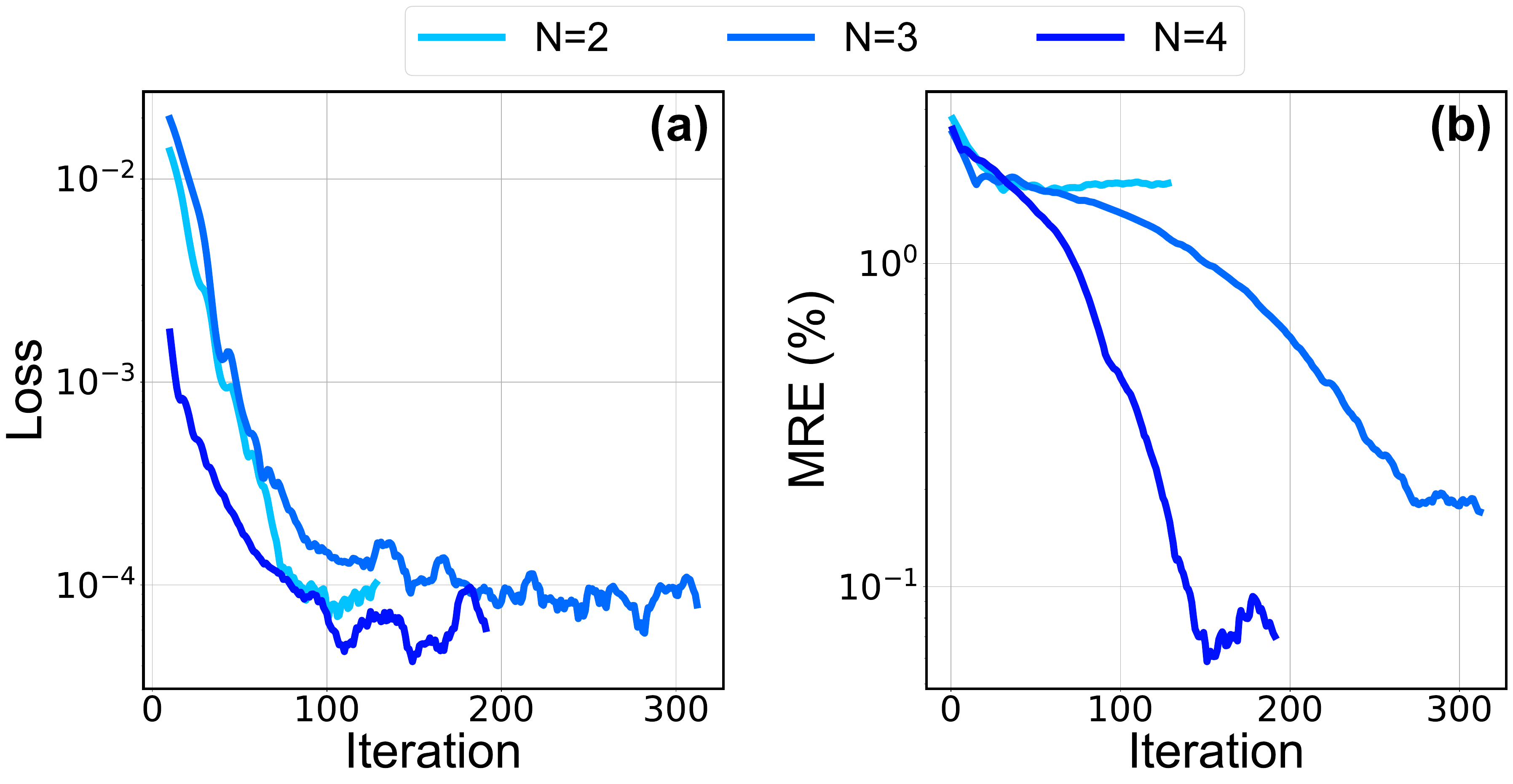}
    \caption{Convergence in a) loss and b) MRE values over iterations for different numbers $N$ of measurements per trial.}
    \label{fig:combined_K_num}
    \vspace{-0.2cm}
\end{figure}

To evaluate the effectiveness of our VLM position selection (VLMSel), we compare its performance against RandSel baselines with $N=3$ receiver positions and VLM initialization.
As shown in Fig.~\ref{fig:my_combined_plot}, our VLMSel method significantly outperforms random selection approaches.
VLMSel achieves faster convergence and a lower MRE (approximately 10 times) compared with RandSel, given conductivities initialized by either VLMInit or RandInit.
% With VLM initialization, RandSel converges at a slower rate than the scheme with VLMSel, resulting in a higher MRE 3.597\%, which costs over three times more iterations to reach a comparable loss and MRE values.
% Also, RandSel \& RandInit and  completely fails to converge to the ground-truth values, maintaining a high MRE around 50\% throughout the process, indicating that random selection of measurement positions is insufficient for accurate material parameter estimation. 
The results also show that when using the VLM for parameter initialization, we can achieve much lower loss values (10 $\sim$ 100 times) and MRE values (10 $\sim$ 100 times) within $1000$ iterations. 
This is due to the closer initialization provided by the VLM to the ground-truth value.
We analyze how the number of receiver positions $N$ affects our framework. 
Fig.~\ref{fig:combined_K_num} illustrates the convergence across different $N$ values.
Using fewer receiver positions (e.g., $N=2$) results in a high MRE with a low accuracy, although the loss converges quickly. This is because the nonlinear system of equations of the RT engine is underdetermined and more measurements are needed.
Increasing $N$ significantly improves accuracy. Further increasing $N$ yields the best accuracy and two times faster convergence compared to the cases of $N=3$ and $4$.
This demonstrates that selecting more informative measurement positions enhances the sensitivity of the loss function to changes in conductivities, leading to more effective updates and better overall performance.
We visualize the selected measurement positions for different methods in the appendix.

% --- End Figure ---

\subsection{Performance Analysis on Scene Configurations}

\begin{table}[t]
    \centering
    \caption{Convergence of Different RT Depth Values $D$ with VLMInit \& VLMSel}
    \label{tab:complexity_analysis_depth}
    \begin{tabular}{c@{\hskip 4pt}c@{\hskip 4pt}c@{\hskip 4pt}c@{\hskip 4pt}c}
    \toprule
    Depth $D$ & MRE (\%) & Time (s) & Iter. & Per Iter. Time (s) \\
    \midrule
    1 & 0.467 & 1370.49 & 409 & 3.35 \\
    2 & 0.504 & 1362.38 & 406 & 3.36 \\
    4 & 0.273 & 997.19 & 296 & 3.37 \\
    6 & 0.070 & 703.07 & 191 & 3.68 \\
    \bottomrule
    \end{tabular}
\end{table}
\begin{table}[t]
    \centering
    \caption{Convergence of Different Numbers $U_{\text{ray}}$ of Rays}
    \label{tab:complexity_analysis_n_ray}
    \begin{tabular}{c@{\hskip 4pt}c@{\hskip 4pt}c@{\hskip 4pt}c@{\hskip 4pt}c}
    \toprule
    $U_{\text{ray}}$ & MRE (\%) & Time (s) & Iter. & Per Iter. Time (s) \\
    \midrule
    3000 & 0.185 & 624.33 & 183 & 3.41 \\
    4000 & 0.160 & 1530.99 & 457 & 3.35 \\
    5000 & 0.070 & 1198.73 & 357 & 3.36 \\
    6000 & 0.032 & 880.81 & 263 & 3.35 \\
    \bottomrule
    \end{tabular}
\end{table}

The impact of ray-tracing depth is shown in Table~\ref{tab:complexity_analysis_depth}. 
Table~\ref{tab:complexity_analysis_depth} summarizes the convergence behavior of our method for different RT depth values~$D$ when using VLMInit and VLMSel. Across all settings, the MRE remains low, demonstrating stable estimation performance. Although the total runtime and number of iterations vary with depth, the per-iteration computation time remains nearly constant (approximately 3.3--3.4\,s for most cases), indicating that each optimization step scales efficiently. Notably, larger depths (e.g., $D=6$) require fewer iterations and yield the lowest MRE, suggesting that increased depth can improve convergence without incurring significant additional per-iteration cost.

Table~\ref{tab:complexity_analysis_n_ray} presents the convergence behavior under different numbers of tracing rays $U_{\text{ray}}$. Increasing $U_{\text{ray}}$ consistently improves the estimation accuracy, as reflected by the decreasing MRE from 0.185 to 0.032. While the total runtime and number of iterations vary across ray counts, the per-iteration computation time remains constant (around 3.3\,s), indicating that the computational cost of each optimization step scales well with the number of rays. Similar to the trend observed for RT depth, a larger number of rays leads to faster convergence in terms of iterations and yields significantly lower MRE, demonstrating the benefit of richer angular sampling without increasing per-iteration overhead.

% fixed $S=1$ and varied $M$ from $5$ to $15$. As shown in Figure~\ref{fig:scalability_plots}(a), the \textbf{average iteration time} scales linearly ($O(M)$), as expected since each material adds a parameter to the backward pass (e.g., $M=5 \to 0.66$s; $M=13 \to 11.54$s). One-time setup costs (Build and Reference Gain) showed negligible dependence on $M$.

% \textbf{Scalability with $N$}: We fixed $M=9$ and varied $S$ from $1$ to $5$, with each setup simulated as an independent scene. Figure~\ref{fig:scalability_plots}(b) reveals that both the average iteration time and the one-time reference gain Time scale in a significant, super-linear fashion with $N$. Doubling $N$ from $2$ to $4$ caused iteration time to increase from $1.94$s (at $N=1$) to $36.15$s (at $N=4$), while total setup time increased from $2.83$s to $15.62$s.

% These results confirm the framework scales linearly with scene complexity ($M$) but highlight a computational bottleneck, as optimization time scales non-linearly with the number of measurement setups ($S$). This suggests our framework is well-suited for high-$M$ scenes but, in its current form, benefits from methods (like our VLM-Select) that minimize the required $N$.

\section{Conclusion}
This work introduced a VLM-guided framework that accelerates and stabilizes multi-material RF parameter estimation within a differentiable RT engine. By parsing scene images to infer material categories and mapping them to ITU-R priors, the method provides informed conductivity initialization; by selecting informative Tx/Rx placements, it increases path diversity and material discriminability. Integrated with NVIDIA Sionna, the approach achieved 2–4× faster convergence and 10–100× lower final error than uniform or random baselines, reaching sub-0.1\% mean relative error with few receivers. These results show that high-level semantic priors from VLMs can effectively guide physics-based optimization for reliable and efficient RF material estimation. Future directions include jointly estimating permittivity and conductivity, closed-loop active measurement with online VLM–DRT co-design, uncertainty-aware planning, robustness to real photographs and partial views, and hardware-in-the-loop validation in real indoor environments.

\section*{Acknowledgments}
This work was supported in part by SUTD Kickstarter Initiative (SKI 2021\_06\_08), in part by NVIDIA Academic Grant Program, and in part by the National Research Foundation, Singapore and Infocomm Media Development Authority under its Communications and Connectivity Bridging Funding Initiative.

\bibliography{aaai2026}
\newpage
\appendix

\onecolumn
\section*{Appendix: Isaac Sim Rendered Scenes, VLM Prompts and Reasoning Process}
We present the detailed VLM prompts and its reasoning process. We also illustrate the Isaac Sim rendered scenes with different number of objects and the VLM selected or randomly selected measurement positions.

\begin{figure*}[!ht]
    \centering
    \subfloat[K=7]
        {\includegraphics[width=0.24\columnwidth]{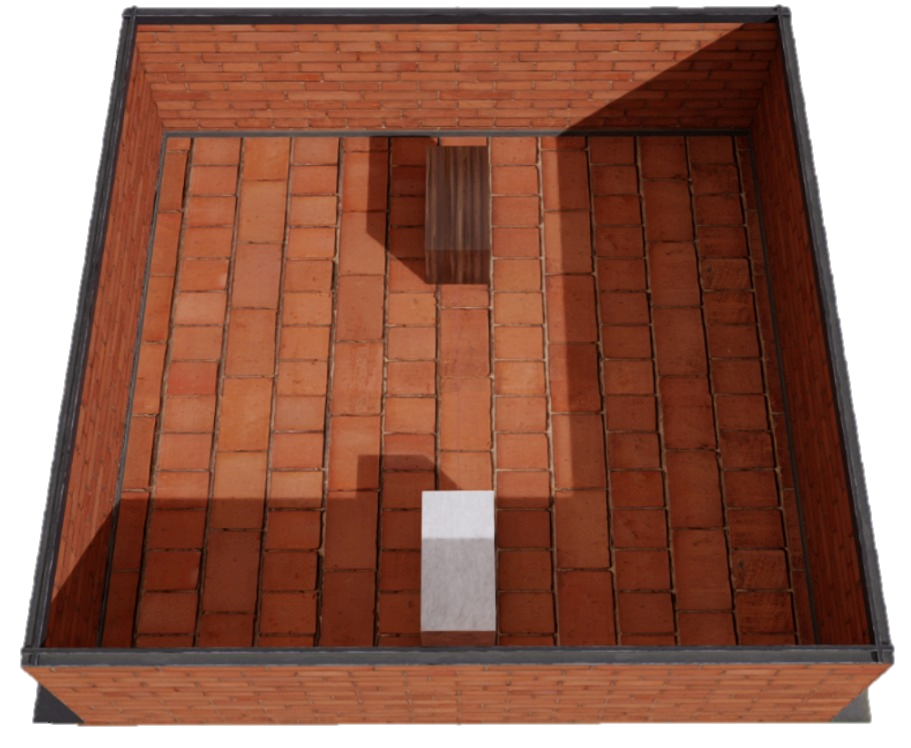}
        \label{fig:scene_K=2}}
    \hfill 
    \subfloat[K=9]
        {\includegraphics[width=0.23\columnwidth]{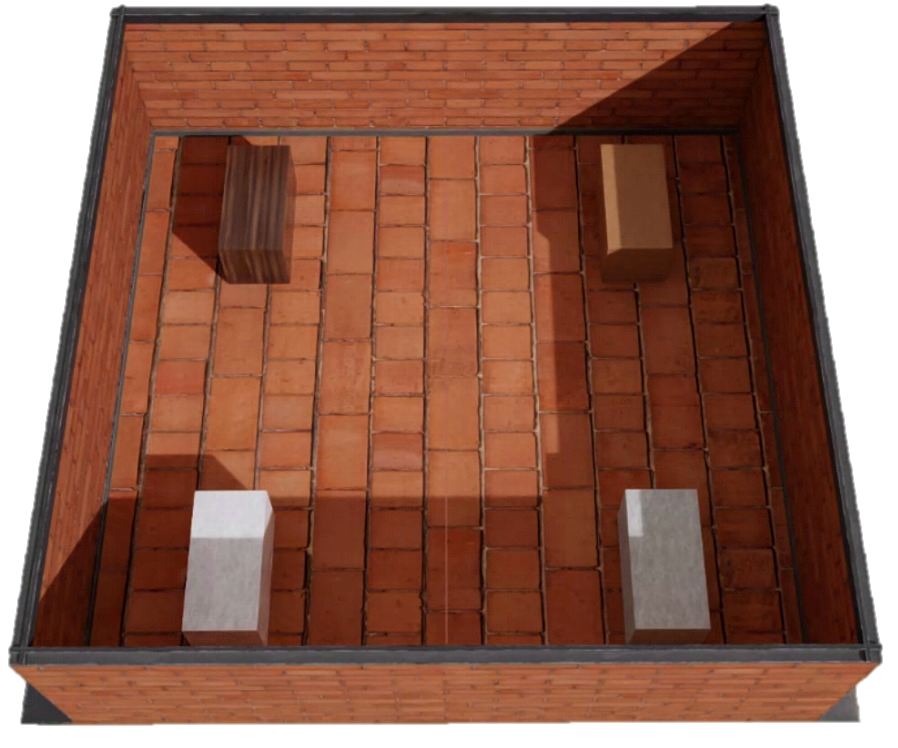}
        \label{fig:scene_K=4}}
    \hfill 
    \subfloat[K=11]
        {\includegraphics[width=0.24\columnwidth]{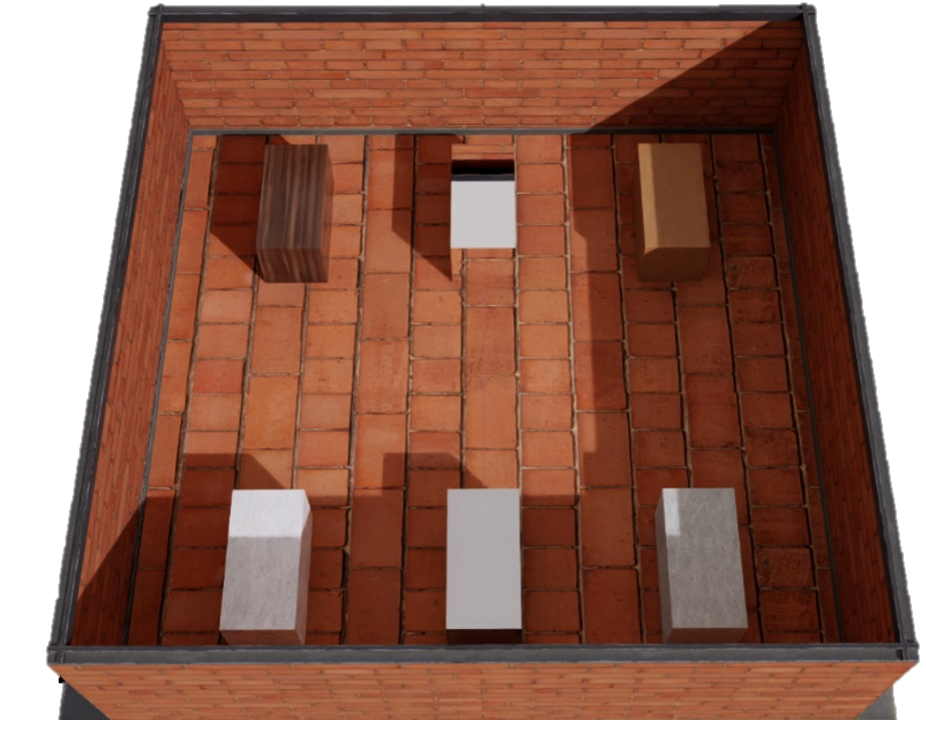}
        \label{fig:scene_K=6}}
    \hfill
    \subfloat[K=13]
        {\includegraphics[width=0.23\columnwidth]{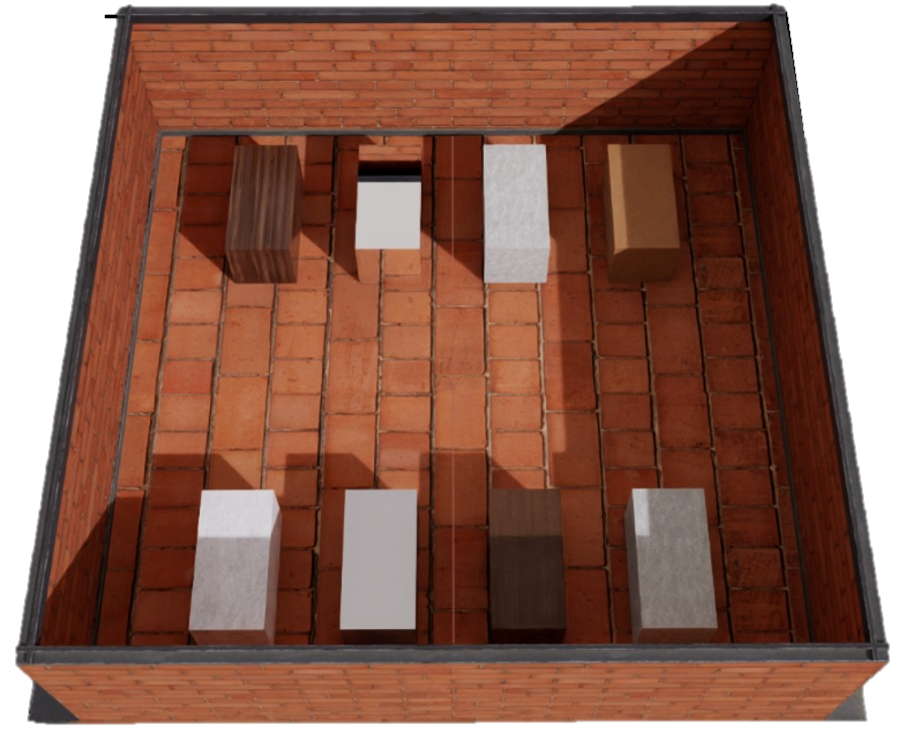}
        \label{fig:scene_K=8}}
    \caption{Isaac Sim scenes with different number of objects $K$. }
    \label{fig:Isaac sim scenes with different num of obj}
\end{figure*}

\begin{figure*}[!ht]
    \centering
    \subfloat[VLMSel (N=4, K=9)]
        {\includegraphics[width=0.45\columnwidth]{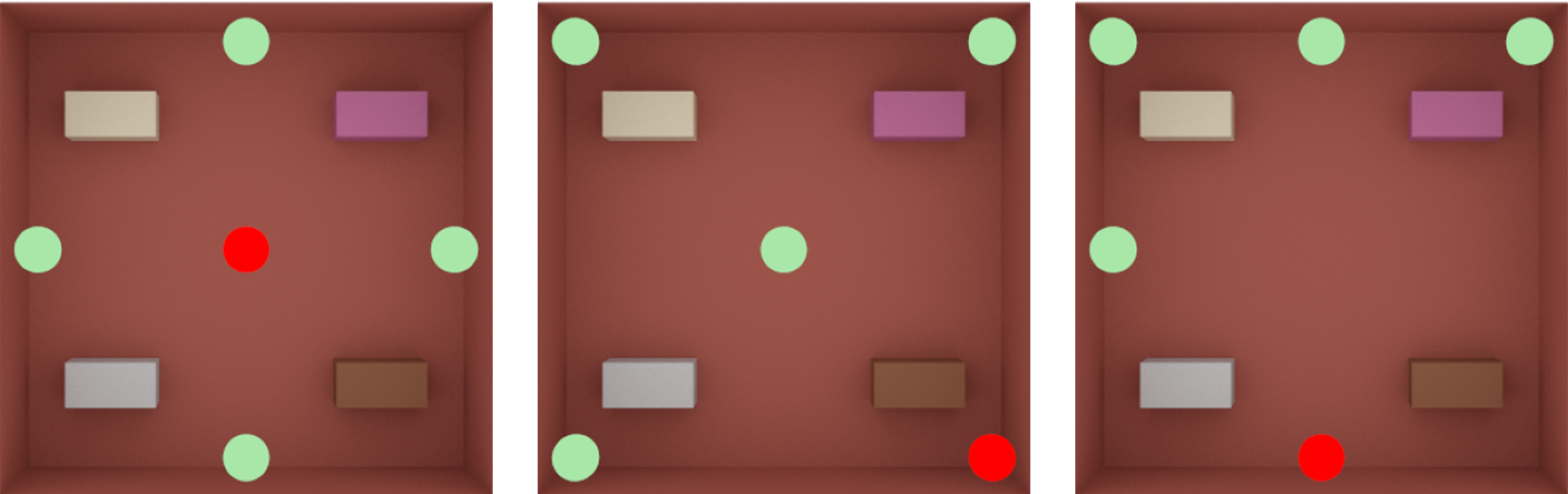}
        \label{fig:pos_vlm_diverse}}
    \hfill 
    \subfloat[RandSel (N=3, K=9)]
        {\includegraphics[width=0.45\columnwidth]{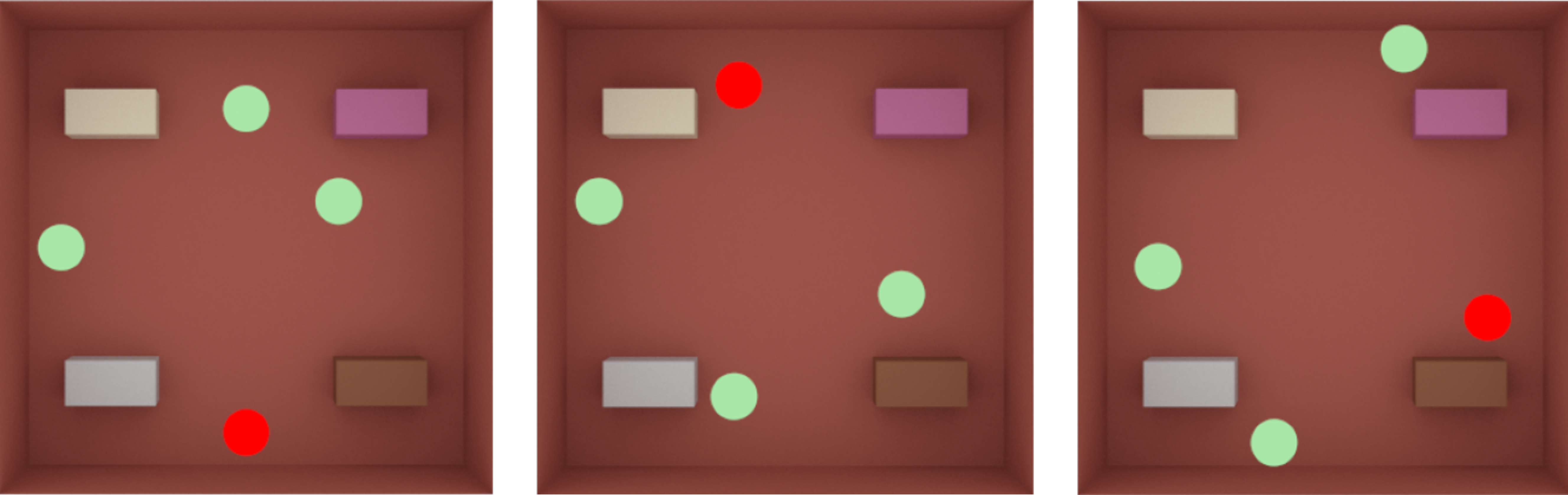}
        \label{fig:pos_random}}
    \hfill 
    \subfloat[VLMSel (N=3, K=9)]
        {\includegraphics[width=0.45\columnwidth]{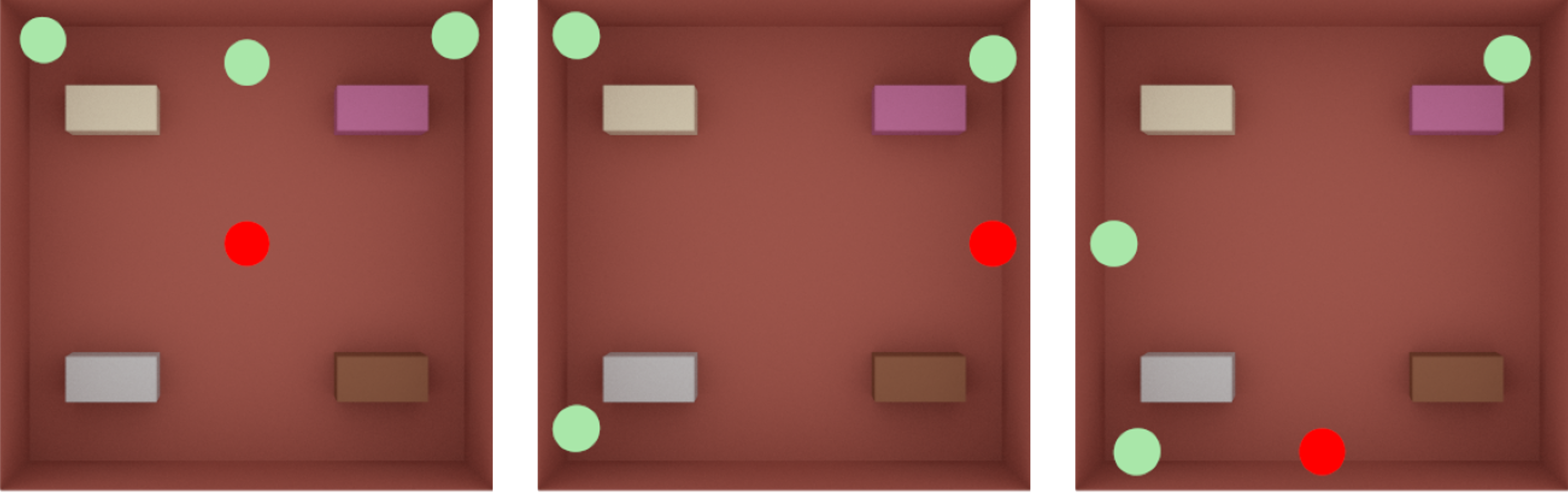}
        \label{fig:pos_vlm_heuristic}}
    \hfill
    \subfloat[VLMSel (N=2, K=9)]
        {\includegraphics[width=0.45\columnwidth]{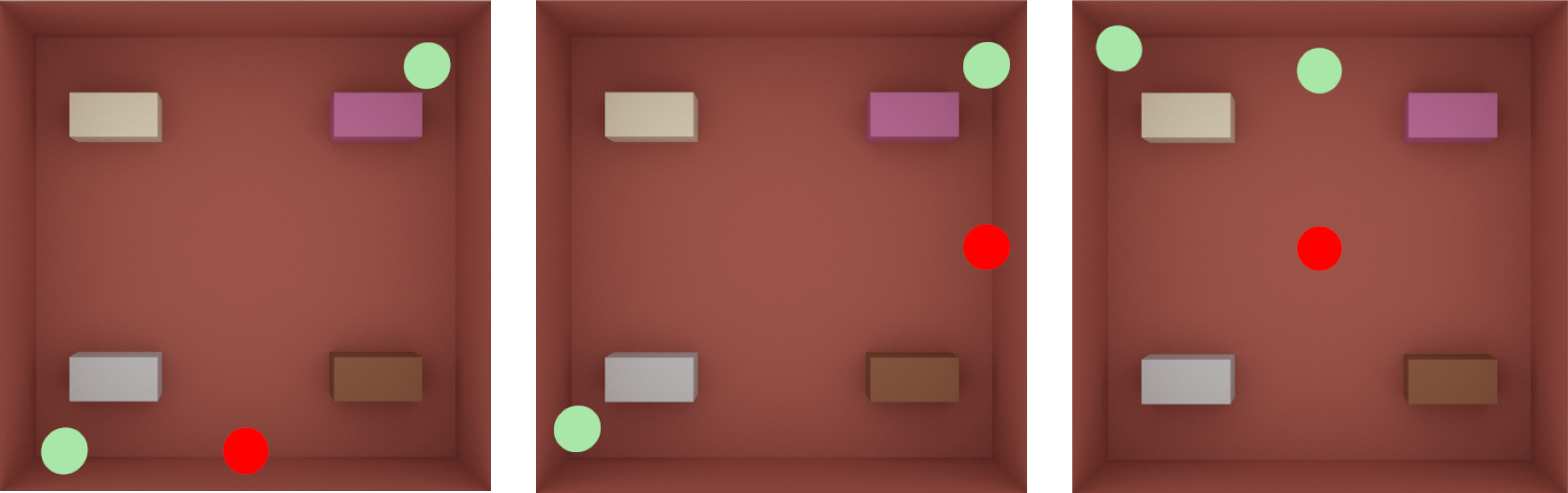}
        \label{fig:pos_vlm_fewest}}
    \caption{
        Measurement positions selected in difference cases. Red and green dots are transmitters and receivers, respectively.
    }
    \label{fig:position_layouts} 
\end{figure*}

% \begin{figure*}[!h]
% \centering
\begin{exmpa}{VLM-Aid RF Material Parameter Initialization}{}
    \small
    Inputs:

    1. Scene Image: A photo of the scene.

    2. ITU-R Material Table: A dictionary mapping standard material names to their (c, d) conductivity parameters.

    ITU\_CONDUCTIVIT\_PARAMS = {

    "Vacuum": (0.0, 0.0),

    "Concrete": (0.0462, 0.7822),

    "Brick": (0.0238, 0.16),

    "Plasterboard": (0.0085, 0.9395),

    "Wood": (0.0047, 1.0718),

    "Glass\_low\_freq": (0.0036, 1.3394),

    "Glass\_high\_freq": (0.0004, 1.658),

    "Ceiling\_board\_low\_freq": (0.0011, 1.0750),

    "Ceiling\_board\_high\_freq": (0.0029, 1.029),

    "Chipboard": (0.0217, 0.78),

    "Plywood": (0.33, 0.0),

    "Marble": (0.0055, 0.9262),

    "Floorboard": (0.0044, 1.3515),

    "Metal": (1e7, 0.0),

    "Very\_dry\_ground": (0.00015, 2.52),

    "Medium\_dry\_ground": (0.035, 1.63),

    "Wet\_ground": (0.15, 1.30),}

    3.Prompt: You are an expert radio frequency (RF) engineer specializing in parameter estimation. Your task is to provide intelligent initial guesses for the frequency-dependent conductivity parameters (c, d) for materials in a given scene, based on visual evidence and standard RF data.

    Task:

    1. Identify all the objects and materials in the scene and for each material, you must return the material names from the ITU-R Material Table.

    2. Map each material name in the scene to their conductivity parameters. If a material is not explicitly listed (e.g., 'Glass'), provide a reasonable default guess based on visual/common sense, or use the value of the most similar material.

    3. Provide the results as a JSON list in such a Schema:
    
    [{"material\_name": "Wall1\_Brick",

        "c": 0.0238,

        "d": 0.16,

        "source": "ITU-R: Brick"},

    {"material\_name": "Wall2\_Brick",
        "c": 0.0238,

        "d": 0.16,

        "source": "ITU-R: Brick"},

    {"material\_name": "Box1\_Wood",
        "c": 0.0047,

        "d": 1.0718,

        "source": "ITU-R: wood"},

    // $\dots$ more materials]
\end{exmpa}

\begin{exmpa}{VLM-Aided Measurement Position Selection}{}
    \small
    Inputs:

    1. Scene Image: An overhead view or a key 3D perspective of the environment.

    2. Geometry XML: The structured XML description of the scene.

    3. Measurement Counts: N (Number of Receivers) = N\_VALUE], M (Number of Tx-Rx measurement configurations) = [M\_VALUE].

    Prompt: You are an intelligent spatial planner optimizing RF measurement campaigns. Your goal is to select Transmitter (Tx) and Receiver (Rx) positions that maximize the diversity and information gain regarding the conductivity parameters of the materials in the scene.

    Task:

    1. Analyze the XML and Image: Identify all distinct material interfaces and complex geometrical features.

    2. Strategy: Select N distinct Tx/Rx positions and form M pairs. Prioritize positions that force the RF path to interact with the different types of materials. For instance, select pairs to sample Non-Line-of-Sight (NLoS) paths or Reflection-dominant paths.

    3. Provide the selected positions as a JSON list. The coordinates must be valid and directly correspond to the XML's coordinate system.

    Output Schema:
    [{"id": "P\_1",
        "type": "Tx",
        "x": 2.5,
        "y": 1.0,
        "z": 1.5,
        "reasoning": "Placed to maximize transmission through the central concrete wall to the back area (NLoS)."},
    {"id": "P\_2",
        "type": "Rx",
        "x": -3.0,
        "y": 5.0,
        "z": 1.5,
        "reasoning": "Paired with P\_1 to measure loss through multiple materials (wood table and brick wall)."},
        
    // ... continue until R positions are defined and S pairs are formed implicitly by the list]

\end{exmpa}

\begin{exmpb}{Reasoning of Measurement Position Selection Using Gemini2.5 Pro}{}
    \small
    Considering User Intent

    I'm currently focused on dissecting the user's initial request. I'm carefully examining the image and the associated XML file to understand the desired goal and the relationship between the visual and textual data. I'm exploring the data and seeking to decipher the key problem the user is trying to solve.

    Interpreting Scene Geometry

    My current focus is on interpreting the scene geometry described in the floor\_wall.xml file. I've successfully identified the floor as a 10x10 plane centered at (0,0,0). I'm now examining the positioning of the four internal boxes to ensure that any coordinate generation avoids collisions. The goal is to establish valid transmitter and receiver locations within the constrained space.

    Refining Spatial Constraints

    I'm now calculating precise boundaries for the safe area. The floor is confirmed to be from -5 to 5 in both X and Y. The walls define the boundaries. The Z-axis runs from 0 to 3. I'm focusing on defining safe zones for TX and RX placement within these constraints, ensuring no collisions with walls.

    Analyzing Spatial Layout

    I'm now carefully examining the scene boundaries. The floor extends from -5 to 5 in both X and Y. The walls are also clearly defined. However, I've realized a potential inconsistency. The XML defines boxes with a small Z-scale (0.2), which conflicts with their representation in the provided image. I will trust the XML data. I will assume the box.ply file is scaled as specified in the XML. I'm now focusing on the positioning of the four boxes within the scene and avoiding collisions.

    Determining Valid Zones

    I'm now determining safe zones for the transmitter and receivers, considering the boxes. Based on the XML, the boxes are relatively small. I need to avoid placement directly on or inside the boxes. My goal is to maximize the distances between the transmitter, receivers, and the surrounding obstacles.

    Generating Safe Locations

    I'm now generating coordinates for the TX and RX groups, focusing on maximizing distances between them and obstacles. I'm using the XML data to define safe areas, accounting for the boxes' small dimensions. The goal is to avoid collisions while ensuring adequate separation.

    Determining Acceptable Locations

    I've determined a safe zone for the TX and RX groups to be within the room, avoiding walls and boxes. I'm taking the box size seriously and treating the XML data as definitive. The goal is to maximize the distances between all devices within the constraints.

    Generating Coordinates

    I'm now generating coordinates for the TX and RX groups, ensuring they're safe from collisions. I'm focusing on the center of the safe zones, and will vary positions around this point. My focus is on separating each group to ensure accurate measurements.

    Finalizing Safe Locations

    I'm now generating final coordinates, ensuring ample separation from all obstacles. I've concluded the most likely scenario is a 1x1 base mesh with a z-scale of 0.2, and will avoid the boxes and walls. My focus is on creating well-separated transmitter and receiver groups.

    Refining Location Strategy

    I'm now generating coordinates for the TX and RX groups, focusing on maximizing distances while adhering to constraints. The XML's small box dimensions make placement easier. I'm focusing on creating well-separated transmitter and receiver groups.

    Validating Safe Positions

    I've rigorously analyzed the scene and the constraints. I'm now generating coordinates for three groups, focusing on maximizing distances from obstacles. My goal is to produce valid TX/RX positions. I'm prioritizing the avoidance of collisions with walls and boxes.

    Validating Location Coordinates

    I'm now generating final coordinates, ensuring ample separation from obstacles. The XML's size for the boxes remains the critical reference point. I will maintain a consistent separation strategy, ensuring that each coordinate group facilitates the measurements. I'm focusing on minimizing the risk of collisions during the raytracing simulation.

    Analyzing Position Constraints

    I'm now focusing on calculating potential coordinate positions. Following that, I will validate the feasibility of each coordinate to ensure it is collision-free. I am still keeping in mind that the primary need is for the three groups of locations, and nothing else.

    Determining Safe Coordinates

    I'm now generating final coordinates, ensuring ample separation from obstacles. The XML's size for the boxes remains the key reference point, and the image is secondary. I will maintain the earlier separation strategy. My current goal is to produce three groups of valid coordinates, with each group containing a single transmitter and eight receivers.
\end{exmpb}

\end{document}